%% file: main.tex
\documentclass[10pt]{article} 
\usepackage[preprint]{tmlr}

\input{math_commands.tex}

\usepackage{hyperref}
\usepackage{url}
\usepackage{booktabs}

\input{extra_packages}
\title{Tighter sparse variational Gaussian processes}

\author{\name Thang D.~Bui \email thang.bui@anu.edu.au \\
      \addr School of Computing\\
      Australian National University
      \AND
      \name Matthew Ashman \email mca39@cam.ac.uk \\
      \addr Department of Engineering \\
      University of Cambridge
      \AND
      \name Richard E.~Turner \email ret26@cam.ac.uk\\
      \addr Department of Engineering \\
      University of Cambridge}

\def\month{MM}  
\def\year{YYYY} 
\def\openreview{\url{https://openreview.net/forum?id=XXXX}} 

\begin{document}

\maketitle

\input{main_text}

\subsubsection*{Acknowledgments}
We thank Martin Jankowiak for pointing out a parameterisation issue in an earlier version of this paper.

\bibliography{references}
\bibliographystyle{tmlr}

\appendix

\input{appendix_text}
\end{document}

%% file: math_commands.tex
\usepackage{amsmath,amsfonts,bm}

\def\eqref#1{equation~\ref{#1}}

\def\1{\bm{1}}

\def\rvm{{\mathbf{m}}}

\def\rmC{{\mathbf{C}}}

\def\rmI{{\mathbf{I}}}

\def\rmM{{\mathbf{M}}}

\def\rmS{{\mathbf{S}}}

\def\rmV{{\mathbf{V}}}

\def\rmX{{\mathbf{X}}}

\def\vzero{{\bm{0}}}

\def\vc{{\bm{c}}}

\def\vf{{\bm{f}}}

\def\vu{{\bm{u}}}
\def\vv{{\bm{v}}}

\def\vx{{\bm{x}}}
\def\vy{{\bm{y}}}
\def\vz{{\bm{z}}}

\DeclareMathAlphabet{\mathsfit}{\encodingdefault}{\sfdefault}{m}{sl}
\SetMathAlphabet{\mathsfit}{bold}{\encodingdefault}{\sfdefault}{bx}{n}

\def\gF{{\mathcal{F}}}

\def\gL{{\mathcal{L}}}

\def\gN{{\mathcal{N}}}
\def\gO{{\mathcal{O}}}

\def\sR{{\mathbb{R}}}



%% file: extra_packages.tex
\usepackage{amsmath,amsfonts,amssymb}
\usepackage{cancel}
\usepackage{cleveref}
\usepackage{mathtools}
\usepackage{multirow}

\newcommand{\kff}{\mathbf{K}_{\mathbf{ff}}}
\newcommand{\kfu}{\mathbf{K}_{\mathbf{fu}}}
\newcommand{\vuv}{[\vu^\intercal, \vv^\intercal]^\intercal}
\newcommand{\kfuv}{\mathbf{K}_{\mathbf{f,uv}}}
\newcommand{\kuvf}{\mathbf{K}_{\mathbf{uv,f}}}
\newcommand{\kfnu}{\mathbf{k}_{f_n\mathbf{u}}}
\newcommand{\kufn}{\mathbf{k}_{\mathbf{u}f_n}}
\newcommand{\kuf}{\mathbf{K}_{\mathbf{uf}}}

\newcommand{\kvv}{\mathbf{K}_{\mathbf{vv}}}
\newcommand{\cvv}{\mathbf{C}_{\mathbf{vv}}}
\newcommand{\cvvinv}{\mathbf{C}_{\mathbf{vv}}^{-1}}
\newcommand{\kuuinv}{\mathbf{K}_{\mathbf{uu}}^{-1}}
\newcommand{\kuvinv}{\mathbf{K}_{\mathbf{uv,uv}}^{-1}}
\newcommand{\qff}{\mathbf{Q}_{\mathbf{ff}}}
\newcommand{\dff}{\mathbf{D}_{\mathbf{ff}}}

\newcommand{\ksu}{\mathbf{k}_{\mathbf{*u}}}
\newcommand{\kus}{\mathbf{k}_{\mathbf{u*}}}
\newcommand{\ksf}{\mathbf{k}_{\mathbf{*f}}}
\newcommand{\kfs}{\mathbf{k}_{\mathbf{f*}}}
\newcommand{\qsf}{\mathbf{Q}_{\mathbf{*f}}}
\newcommand{\qfs}{\mathbf{Q}_{\mathbf{f*}}}

\newcommand{\kss}{\mathbf{k}_{\mathbf{**}}}
\newcommand{\kvu}{\mathbf{K}_{\mathbf{vu}}}
\newcommand{\kuv}{\mathbf{K}_{\mathbf{uv}}}

\newcommand{\gp}{\mathcal{GP}}
\newcommand{\kl}{\mathrm{KL}}
\newcommand{\tr}{\mathrm{trace}}
\newcommand{\diag}{\mathrm{diag}}

%% file: main_text.tex
\begin{abstract}
Sparse variational Gaussian process (GP) approximations based on inducing points have become the de facto standard for scaling GPs to large datasets, owing to their theoretical elegance, computational efficiency, and ease of implementation. This paper introduces a provably tighter variational approximation by relaxing the standard assumption that the conditional approximate posterior given the inducing points must match that in the prior. The key innovation is to modify the conditional posterior to have smaller variances than that of the prior at the training points. We derive the collapsed bound for the regression case, describe how to use the proposed approximation in large data settings, and discuss its application to handle orthogonally structured inducing points and GP latent variable models. Extensive experiments on regression benchmarks, classification, and latent variable models demonstrate that the proposed approximation consistently matches or outperforms standard sparse variational GPs while maintaining the same computational cost. An implementation will be made available in all popular GP packages.
\end{abstract}


\section{Introduction}
\label{sec:intro}
Gaussian processes (GPs) \citep{rasmussen2006gaussian} provide a powerful framework for modelling probability distributions over functions, offering principled uncertainty quantification and ease of use. Their flexibility in encoding domain knowledge---such as smoothness, peridocity, or domain-specific structure---has led to widespread adoption across scientific and engineering applications. Exact inference in GP models poses significant computational challenges, requiring $\mathcal{O}(N^3)$ time and $\mathcal{O}(N^2)$ space complexity for $N$ observations. A suite of approximations have been developed to address these limitations. Most notably, sparse variational Gaussian processes  \citep[SVGP;][]{titsias2009variational,hensman2015scalable,matthews2016sparse} address the poor computational complexity through the use of an approximate posterior distribution parameterised by a small set of \emph{inducing points}.

The standard SVGP framework employs a structured variational approximation that factorises the posterior distribution over the unknown function $f$ into two components: $q(f) = p(f | \vu) q(\vu)$. Here, $p(f | \vu)$ represents the GP prior distribution conditioned on the function values at inducing locations $\vz$, $\vu = f(\vz)$. The second term, $q(\vu)$, is modelled as a multivariate Gaussian distribution. Improved variational approximations have been developed---such as SOLVE-GP \citep{shi2020sparse}---which use more sophisticated distributions for $q(\vu)$.

This paper introduces a novel approach to improving SVGP approximations by modifying the conditional GP prior distribution at observed inputs, rather than focusing solely on the inducing point distribution. For Gaussian likelihoods, our approach yields a new and improved collapsed lower bound on the log marginal likelihood that involves \emph{no additional variational parameters}. Furthermore, we show how the uncollapsed form of our bound facilitates the use of stochastic mini-batch optimisation and extends naturally to non-Gaussian likelihoods through a single additional variational parameter. We demonstrate the versatility of our method by integrating it with SOLVE-GP and extending it to sparse variational approximations in the GP latent variable model \citep[GPLVM;][]{lawrence05, damianou16a}. Our results demonstrate that by targeting our improved lower bound, our approach consistently improves the predictive performance and log marginal likelihood estimates across a range of regression, classification, and latent variable modelling tasks.

\section{Background}
\label{sec:background}
This section provides a concise introduction to pseudo-point based sparse variational Gaussian processes \citep[SVGP;][]{titsias2009variational,hensman2015scalable,matthews2016sparse}. Consider GP regression with Gaussian observation noise:
\begin{align}
    p(f | \gamma) &= \gp(f; 0, k_\gamma),\\
    p(\vy | f, \vx, \sigma^2) &= \gN(\vy; f(\vx), \sigma^2\rmI),
\end{align}
where $\vx \in \sR^{N\times D}$ and $\vy \in \sR^{N}$ are the training inputs and corresponding noisy outputs, $f$ denotes the unknown function mapping from input to output, $k_\gamma$ is the covariance function governed by hyperparameters $\gamma$, and $\sigma^2$ is the observation noise. These hyperparameters, denoted collectively as $\theta$, can be found by maximising the log marginal likelihood:
\begin{align}
    \gL(\theta) = \vc - \frac{1}{2} \vy^\intercal (\kff + \sigma^2\rmI)^{-1} \vy - \frac{1}{2}\log |\kff + \sigma^2\rmI|,
\end{align}
where $\kff$ is the covariance between training function values $\vf  = f(\vx)$. 
This objective takes $\gO(N^3)$ to compute and is thus computationally prohibitive for large $N$. To sidestep this, we use an approximate posterior judiciously parameterised by a small set of pseudo-points or inducing points as follows:
\begin{align}
    q(f) = p(f_{\neq \vf, \vu} | \vf, \vu) p(\vf | \vu) q(\vu), \label{eq:titsias_q}  
\end{align}
where $\vu = f(\vz) \in \sR^{M}$ and $\vz \in \sR^{M \times D}$ are the inducing outputs and inputs, respectively, and $M \ll N$. The conditional $q(f_{\neq \vu} | \vu)$ in the approximate posterior is chosen to match that in the prior, leading to the following variational objective,
\begin{align}
    \gF_0(q(\vu), \theta) 
        &= \left \langle \log \frac {p(f) p(\vy | f, \vx) } {q(f)} \right \rangle_{q(f)} = \left \langle \frac {\log \cancel{p(f_{\neq \vf, \vu} | \vf, \vu)} \cancel{p(\vf | \vu)} p(\vu) p(\vy | f, \vx) } {\cancel{p(f_{\neq \vf, \vu} | \vf, \vu)} \cancel{p(\vf | \vu)} q(\vu)} \right \rangle_{q(f)} \nonumber \\
        &= - \kl [q(\vu) || p(\vu)] + \sum_{n} \int_{\vu, f(x_n)} q(\vu) p(f({x_n}) | \vu) \log p(y_n | f({x_n})). \label{eq:titsias_bound}
\end{align}
\cite{titsias2009variational} showed that when the likelihood is Gaussian, an analytic optimal form for $q(\vu)$ can be found, $q(\vu) \propto p(\vu) \gN(\vy; \kfu\kuuinv\vu, \sigma^2\rmI)$, and that a collapsed bound is also analytically available,
\begin{align}
    \gF_1(\theta) &= \vc - \frac{1}{2} \vy^\intercal (\qff + \sigma^2\rmI)^{-1} \vy - \frac{1}{2} \log |\qff + \sigma^2\rmI| - \frac{1}{2\sigma^2} \tr (\dff), \label{titsias_collapsed}
\end{align}
where $\qff = \kfu\kuuinv\kuf$ and $\dff = \kff - \qff$. Crucially, the bound above can be computed in $\gO(NM^2)$. The non-collapsed bound in \cref{eq:titsias_bound} is amenable to non-Gaussian likelihoods and data mini-batch settings \citep[see e.g.,][]{hensman2015scalable}, further reducing the training computational complexity to $\gO(BM^2 + M^3)$ where $B$ is the mini-batch size. Due to this small complexity and the ease of implementation, the above variational approach has arguably become the go-to sparse approximation in the GP literature. In this work, we will revisit its core assumption of matching prior and posterior conditionals and show that relaxing this assumption results in a tighter and more performant approximation.

\section{A tighter variational approximation}
\label{sec:tighter_approx}
The variational approximation in \cref{eq:titsias_q} is chosen such that the conditional $q(\vf | \vu)$ identically matches the prior conditional $p(\vf | \vu)$. Instead, we propose using the following variational posterior,
\begin{align}
    q(f) = p(f_{\neq \vf, \vu} | \vf, \vu) q(\vf | \vu) q(\vu), \label{eq:tighter_q}
\end{align}
where $q(\vf | \vu) = \gN(\vf; \kfu\kuuinv\vu; \dff^{1/2}\rmM \dff^{\top/2})$, $\rmM$ is a diagonal matrix, $\rmM = \diag([m_1, m_2, \dots, m_N])$ and $m_n > 0$. Note that the mean of the prior conditional $p(\vf | \vu) = \gN(\vf; \kfu\kuuinv\vu; \dff)$ is retained in $q(\vf | \vu)$. The resulting variational bound is,
\begin{align}
    \gF_2(q(\vu), \theta, \rmM) 
        &= \left \langle \frac {\log \cancel{p(f_{\neq \vf, \vu} | \vf, \vu)} p(\vf | \vu) p(\vu) p(\vy | f, \vx) } {\cancel{p(f_{\neq \vf, \vu} | \vf, \vu)} q(\vf | \vu) q(\vu)} \right \rangle_{q(f)} \nonumber \\
        &= - \kl [q(\vu) || p(\vu)] - \int_\vu q(\vu) \kl [q(\vf | \vu) || p(\vf | \vu)] \nonumber \\ & \qquad +  \sum_{n} \int_{\vu, f(x_n)} q(\vu) q(f({x_n}) | \vu) \log p(y_n | f({x_n})). \label{eq:tighter_uncollapsed}
\end{align}
Due to the structure of the variational distribution, the middle term can be simplified to,
\begin{align}
    - \int_\vu q(\vu) \kl [q(\vf | \vu) || p(\vf | \vu)]
    = - \frac{1}{2} \tr(\rmM) + \frac{1}{2}\log |\rmM| + \frac{N}{2} = \frac{1}{2}\sum_n [1 + \log(m_n) - m_n] \nonumber
\end{align}
\paragraph{Collapsed bound and optimal $\rmM$} In the regression case, similar to the Titsias' bound, we can obtain the optimal form for $q(\vu) \propto p(\vu) \gN(\vy; \kfu\kuuinv\vu, \sigma^2\rmI)$, leading to the following collapsed bound,
\begin{align}
    \gF_3(\theta, \rmM) 
    &=\vc - \frac{1}{2} \vy^\intercal (\qff + \sigma^2\rmI)^{-1} \vy - \frac{1}{2} \log |\qff + \sigma^2\rmI| - \frac{1}{2} \sum_n \left[\frac{m_n d_n}{\sigma^2} - 1 - \log(m_n) + m_n\right] \nonumber
\end{align}
Setting the partial derivatives of $\gF_3(\theta, \rmM)$ wrt $m_n$ to 0, we arrive at $\color{blue}{m_n = \frac{\sigma^2}{d_n + \sigma^2}}$ and the following bound,
\begin{align}
    \color{blue}{\gF_4(\theta) =\vc - \frac{1}{2} \vy^\intercal (\qff + \sigma^2\rmI)^{-1} \vy - \frac{1}{2} \log |\qff + \sigma^2\rmI| - \frac{1}{2} \sum_n \log \left( 1+\frac{d_n}{\sigma^2} \right)}, \label{eq:tighter_collapsed}
\end{align}
where $d_n$ is the $n$-th element in the diagonal of $\dff$, $d_n = {k}_{f_nf_n} - \kfnu\kuuinv\kufn$.
\paragraph{Comparison to Titsias' bound} When $\rmM$ is the identity matrix, that is $m_n = 1$ $\forall n$, the approximation in \cref{eq:tighter_q} become the Titsias' variational approximation in \cref{eq:titsias_q} and the bound in $\gF_3(\theta)$ becomes the Titsias' bound $\gF_1(\theta)$ in \cref{titsias_collapsed}. We note that that $F_4(\theta)$ is tighter than $\gF_1(\theta)$ due to the inequality $\log(1 + x) < x$ for all $x > -1$. Our solution improves upon the solution to the Titsias' bound by allowing the marginals of the conditional approximate posterior, $q(f(x_n) | \vu)$, to have smaller variances than that of the conditional prior, since the optimal $m_n = \frac{\sigma^2}{d_n + \sigma^2} < 1$. Intuitively, this reduces the strength of the coupling between $q(\vu)$ and $q(\vf)$, enabling $q(\vf)$ to more freely model the data whilst allowing $q(f)$ to be close to the prior elsewhere.

It is also worth noting that the middle term of our bound is always non-positive. One might think that adding this term to the bound would give a poorer approximation, yet, the improvement in the expected log-likelihood (due to the smaller predictive variances at the training points---see predictions below) can yield a larger improvement to counteract.

\paragraph{Stochastic mini-batch settings}
The new bound can also handle data mini-batching, yielding an \emph{unbiased} estimator of the uncollapsed bound in \cref{eq:tighter_uncollapsed} as follows,
\begin{align}
    \gF_2(q(\vu), \theta, \rmM) &\approx - \kl [q(\vu) || p(\vu)] + \frac{N}{2B}\sum_b [1 + \log(m_b) - m_b] \nonumber \\ &\qquad\qquad + \frac{N}{B} \sum_{B} \int_{\vu, f(x_b)} q(\vu) q(f({x_b}) | \vu) \log p(y_b | f({x_b})).
\end{align}

\paragraph{Non-Gaussian likelihoods and $m_n$ parameterisation}
One can parameterise $m_n$'s to satisfy their positive constraint and optimise them directly at the cost of having $N$ extra parameters. However, the optimal form for $m_n$ in the Gaussian likelihood setting suggests a more efficient parameterisation $m_n = \beta / (d_n + \beta)$ with $\beta > 0$ shared across all data points. We will use the latter parameterisation for all of our experiments. 

\paragraph{Predictions}
The predictive mean and variance of the predictive distribution at a test input $x_*$ are
\begin{align}
\small
    m_* &= \ksu \kuuinv \rvm_\vu, \\
    v_* &= \kss - \ksu \kuuinv \kus + \ksu \kuuinv \rmS_\vu \kuuinv \kus - (\ksf - \qsf) \rmV_{\mathbf{ff}} (\kfs- \qfs), \label{eq:pred_var}
\end{align}
where $\rmV_\mathbf{ff} = \dff^{-\top/2} (\rmI - \rmM) \dff^{-1/2}$. Note that (i) we can compute the predictive mean at the same cost as previous sparse approximations, and (ii) the predictive variance at a training point can be approximated by $v_n = m_n d_n + \mathbf{k}_{f_n\vu} \kuuinv \rmS_\vu \kuuinv \mathbf{k}_{\vu f_n}$. More generally, the variance at a new input that is not a training or inducing input is expensive due to the presence of $\dff$ in the last term. One path to address this could be to approximate $\dff$ by its diagonal matrix or to use only a subset of training points for this computation. However, we find that simply ignoring the last term at test time does not impact the predictive performance while substantially reducing the prediction cost (see \cref{sec:ablation_prediction}).

\paragraph{Connections to existing bounds}

We can use the log-sum inequality\footnote{For non-negative numbers $a_1, a_2, \ldots, a_n$ and $b_1, b_2, \ldots, b_n$, $\sum_{i=1}^{n} a_i \log \frac{a_i}{b_i} \geq \left(\sum_{i=1}^{n} a_i \right) \log \frac{\sum_{i=1}^{n} a_i}{\sum_{i=1}^{n} b_i}$ with equality iff $a_i/b_i = \text{constant}$.} to bound the last term of our collapsed bound:
\begin{align}
    \sum_{n=1}^N \log \left( 1 + \frac{d_n}{\sigma^2} \right) \leq N \log \frac{ \left[\sum_{n=1}^N \left(1 + \frac{d_n}{\sigma^2}\right)\right]} {N} = N \log \left[1 + \frac{\tr (\kff - \qff)}{N\sigma^2}\right].
\end{align}
Thus a looser collapsed bound can be obtained:
\begin{align}
    \gF_5(\theta) =\vc - \frac{1}{2} \vy^\intercal (\qff + \sigma^2\rmI)^{-1} \vy - \frac{1}{2} \log |\qff + \sigma^2\rmI| - \frac{N}{2} \log \left( 1+\frac{\tr(\kff - \qff)}{N\sigma^2} \right). \nonumber
\end{align}
This bound was derived by \citet{artemev2021tighter} based on bounds of the quadratic and log-determinant terms in the exact log marginal likelihood. This is also tighter than the Titsias' bound, that is $\gF_4(\theta) \geq \gF_5(\theta) \geq \gF_1(\theta)$.

One can also view the proposed variational approximation as an instance of the sparse orthogonal approach of \citet{shi2020sparse} in which there are two sets of inducing points $\vu$ and $\vv$ with $\vv \vcentcolon= \vf$, $\rvm_\vv \vcentcolon= \vzero$ and $\rmS_\vv \vcentcolon= \dff^{1/2}\rmM \dff^{\top/2}$. However, this view does not suggest new insights or potential improvements. We will next discuss how to use the proposed variational approximation to improve the sparse orthogonal approach and in the latent variable settings.


\section{Application to sparse orthogonal variational GPs}
The sparse orthogonal approach (SOLVEGP) of \citet{shi2020sparse} can be viewed as a structured approximation with two sets of pseudo-points $\vu$ and $\vv$,
\begin{align}
    q(f) &= p(f_{\neq \vf, \vu, \vv} | \vf, \vu, \vv) p(\vf | \vu, \vv) q(\vu, \vv), \nonumber \\
    q(\vu, \vv) &= \gN(\vu; \rvm_\vu, \rmS_\vu) \gN(\vv; \kvu \kuuinv \vu + \rvm_\vv, \rmS_\vv) \nonumber \\ &= \gN\left( \begin{bmatrix} \vu \\ \vv \end{bmatrix} ; \begin{bmatrix} \rvm_\vu \\ \kvu\kuuinv \rvm_\vu + \rvm_\vv \end{bmatrix} , \begin{bmatrix} \rmS_\vu & \rmS_\vu \kuuinv \kuv \\ \kvu \kuuinv \rmS_\vu & \rmS_\vv + \kvu\kuuinv \rmS_\vu \kuuinv\kuv \end{bmatrix} \right),\nonumber
\end{align}
where $(\rvm_\vu, \rmS_\vu)$ and $(\rvm_\vv, \rmS_\vv)$ are the mean and covariance variational parameters. This approximation brings computational benefits over naively using a single set of pseudo-points with cardinality $M = M_\vu + M_\vv$ while matching the latter's performance. We will show that the same trick used for sparse variational GPs---relaxing the conditional matching assumption $q(\vf | \vu, \vv) = p(\vf | \vu, \vv)$--- can improve SOLVEGP. In particular, similar to SVGP, we will use $q(\vf | \vu, \vv) = \gN(\vf; \kfuv\kuvinv\vuv; \dff^{1/2} \rmM \dff^{\top/2})$ where $\dff = \kff - \kfuv\kuvinv\kuvf$, $\rmM$ is a diagonal matrix, $\rmM = \diag([m_1, m_2, \dots, m_N])$ and $m_n > 0$. The resulting variational bound is 

\begin{align}
    \gF_6(q(\vu, \vv), \theta, \rmM) 
        &= \left \langle \frac {\log \cancel{p(f_{\neq \vf, \vu, \vv} | \vf, \vu, \vv)} p(\vf | \vu, \vv) p(\vu, \vv) p(\vy | f, \vx) } {\cancel{p(f_{\neq \vf, \vu, \vv} | \vf, \vu, \vv)} q(\vf | \vu, \vv) q(\vu, \vv)} \right \rangle_{q(f)} \nonumber \\
        &= - \kl [q(\vu) || p(\vu)] - \kl [\tilde{q}(\vv) || \tilde{p}(\vv)] + \frac{1}{2}\sum_n [1 + \log(m_n) - m_n] \nonumber \\ & \qquad \qquad +  \sum_{n} \int_{\vu,\vv,f(x_n)} q(\vu,\vv) q(f({x_n}) | \vu, \vv) \log p(y_n | f({x_n})). \label{eq:tighter_solvegp},
\end{align}
where $\tilde{q}(\vv) = \gN(\vv; \rvm_\vv, \rmS_\vv)$, $\tilde{p}(\vv) = \gN(\vv; \vzero, \cvv)$, and $\cvv = \kvv - \kvu\kuuinv\kuv$. Note that the predictive distribution at a training point can be approximated efficiently, $q(f(x_n)) \approx \gN(f(x_n); m_n, v_n)$ with 
\begin{align}
    m_n &= \mathbf{k}_{f_n\vu} \kuuinv \rvm_\vu + \mathbf{c}_{f_n\vv} \cvvinv \rvm_\vu, \nonumber \\
    v_n &= m_n (\mathbf{c}_{f_nf_n} - \mathbf{c}_{f_n\vv} \cvvinv \mathbf{c}_{\vv f_n})  + \ksu \kuuinv \rmS_\vu \kuuinv \kus + \mathbf{c}_{f_n\vv} \cvvinv \rmS_\vv \cvvinv \mathbf{c}_{f_n\vv} \nonumber
\end{align}
where $\mathbf{c}_{ab} = \mathbf{k}_{ab} - \mathbf{k}_{a\vu} \kuuinv \mathbf{k}_{\vu b}$. Similar to SVGP, the predictive variance at a new test point is expensive due to the dependence on all training points. However, similar to the tighter approximation in \cref{sec:tighter_approx}, we found that simply ignoring this difficult term works well in practice.

\section{Application to Bayesian GP latent variable models}
\label{sec:lvm}
Consider a GP latent variable model \citep[GPLVM;][]{lawrence05} with Gaussian observation noise:
\begin{gather}
    p(\vx) = \gN(\vx; 0, \rmI), \nonumber\\
    p(f | \gamma) = \gp(f; 0, k_{\gamma}), \nonumber\\
    p(\vy | f, \vx) = \gN(\vy; f(\vx), \sigma^2\rmI). \nonumber
\end{gather}
Both the posterior $p(f | \vy)$ and marginal likelihood $p(\vy)$ are intractable. Instead, we introduce an approximate posterior of the following form:
\begin{gather}
    q(f, \vx) = q(\vx) p(f_{\neq \vf, \vu} | \vf, \vu) q(\vf | \vu, \vx) q(\vu) \nonumber\\
    q(\vf | \vu, \vx) = \gN(\vf; \kfu \kuuinv \vu, \dff^{1/2} \rmM(\vx) \dff^{\top/2}),\nonumber
\end{gather}
where $\rmM(\vx) = \diag([m_1(x_1), m_2(x_2), \dots, m_N(x_N)])$ and $m_n(x_n) > 0$. Note that $q(\vf | \vu, \vx)$ depends on $\vx$ through $\kfu$, $\dff$ and $\rmM(\vx)$, and that when $\rmM$ is the identity matrix, that is $m_n = 1$, we obtain the variational approximation of \cite{damianou16a}. We can bound the log marginal likelihood as
\begin{align}
    \gF(q(f, \vx), \theta) &= -\kl [q(\vx) \| p(\vx)] -\kl [q(\vu) \| p(\vu)] + \frac{1}{2} \sum_n \left \langle 1 + \log(m(x_n)) - m(x_n)\right\rangle_{q(x_n)} \nonumber\\
    & \qquad\qquad + \sum_n \int_{\vu, x_n, f(x_n)} q(x_n)q(\vu)q(f(x_n) | x_n, \vu) \log p(y_n | f(x_n)). \nonumber
\end{align}
We can obtain the collapsed bound by noting that the optimal form for $q(\vu)$ is given by
\begin{equation}
    q(\vu) \propto p(\vu) \exp\left(\langle \log \gN(\vy; \kfu \kuuinv \vu, \sigma^2 \rmI) \rangle_{q(\vx)} \right). \nonumber
\end{equation}
Note also that
\begin{equation}
    \int_{\vu, \vf} q(\vu) q(\vf | \vu) \log p(\vy | \vf) = \int_{\vu} q(\vu) \log \gN(\vy; \kfu \kuuinv \vu, \sigma^2 \rmI) - \sum_n \frac{m_n(x_n) d_n}{2\sigma^2}. \nonumber
\end{equation}
Together with Jensen's inequality, we arrive at the collapsed bound
\begin{align}
    \gF(q(\vx))
    &= -\kl [q(\vx) \| p(\vx)] - \frac{1}{2} \sum_n \left \langle \frac{m_n(x_n) d_n}{2\sigma^2} - 1 - \log m_n(x_n) + m_n(x_n)\right \rangle_{q(x_n)} \nonumber \\ & \qquad + \log \left( \int_{\vu} e^{\langle \log \gN(\vy; \kfu \kuuinv \vu, \sigma^2 \rmI) \rangle_{q(\vx)}} p(\vu)\right). \nonumber
\end{align}
Setting derivatives w.r.t.\ $m_n(x)$ to 0 gives
\begin{align}
    \langle m_n(x_n) \rangle_{q(x_n)} = \left\langle \frac{\sigma^2}{d_n + \sigma^2} \right\rangle_{q(x_n)} \nonumber
\end{align}
which is satisfied by $m_n(x_n) = \frac{\sigma^2}{d_n + \sigma^2}$ or $m_n(x_n) = \left\langle \frac{\sigma^2}{d_n + \sigma^2} \right\rangle_{q(x_n)}$. The former is easier to implement as we do not need to (approximately) integrate out $x_n$ to find $m_n$.

\begin{figure}[t]
    \centering
    \includegraphics[width=\linewidth]{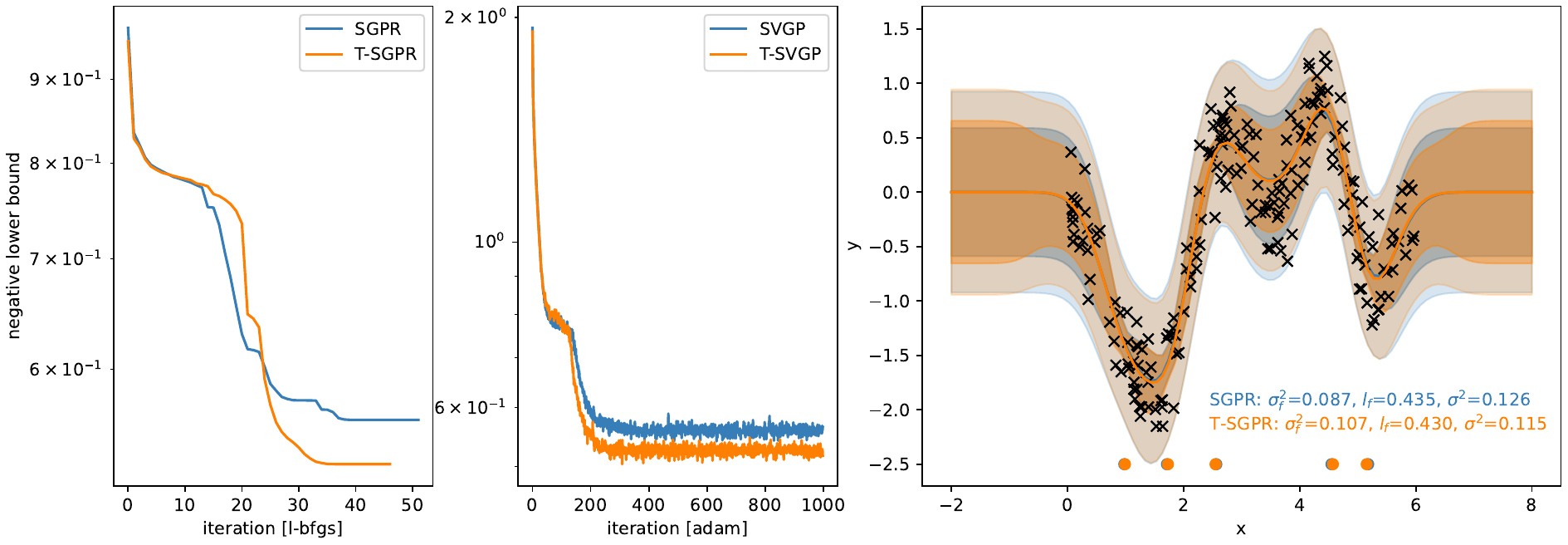}
    \caption{Left and middle: Optimisation traces for SGPR, T-SGPR, SVGP and T-SVGP on the Snelson dataset with 5 inducing points. Right: Predictive means and uncertainties. The stronger shade is for noiseless predictions.}
    \label{fig:snelson}
\end{figure}

\section{Experimental results}
We validate the utility of the proposed variational posterior in a suite of experimental settings. We switch the variational objective with the proposed approximation in each setting, keep all other configurations unchanged, and measure the two's predictive performance. Implementations based on GPytorch and GPflow will be released.

\subsection{Toy 1-D regression}
To build intuition about the proposed method's behaviour, we first evaluate it on a 1-D regression problem used by \cite{snelson2005sparse}. We compare (i) Titsias's collapsed bound in \cref{eq:titsias_bound} [SGPR] with the proposed collapsed bound in \cref{eq:tighter_collapsed} [T-SGPR], and (ii) Titsias's uncollapsed bound [SVGP] with the proposed uncollapsed bound in \cref{eq:tighter_uncollapsed} [T-SVGP]. \Cref{fig:snelson} illustrates the optimisation trajectories of these methods and the final fits for both SGPR and T-SGPR using five inducing points. The final values for both uncollapsed and collapsed versions of the proposed bound appear tighter than that of the Titsias' bound in practice. The learned hyperparameters reveal that T-SGPR prefers smaller observation noise (0.115) and larger kernel variance (0.107) compared to that of SGPR (0.126 and 0.087, respectively).

\subsection{Efficient predictive variances}
A key practical consideration is the computational cost of predictive variance in \cref{eq:pred_var}. The exact computation requires $\dff^{-1}$ which scales poorly with the training set size. We evaluate a simplified variant that omits the term that involves $\dff$, the last term in \cref{eq:pred_var}, and compare it to the exact variance calculation across three small benchmark datasets: wine, solar, and pumadyn32nm.
Table 1 presents a detailed comparison between the exact and approximate versions. We can see a pattern across all datasets: the simplified variant consistently matches the full model's performance while offering substantial computational savings. For this reason, we will be using the simplified version for all remaining experiments. The improvements in predictive performance in \cref{sec:exp_reg,sec:exp_mnist} are therefore solely due to better estimation of the hyperparameters and a different $q(\vu)$.
\label{sec:ablation_prediction}

\begin{table}[!ht]
\small
\centering
\begin{tabular}{lcc ccc}
\toprule
Dataset & N/D & \cref{eq:pred_var} & RMSE & Log-likelihood & Time (s) \\
\midrule
\multirow{2}{*}{wine} &
\multirow{2}{*}{1599/11}
 & w. last term & 0.47 $\pm$ 0.01 & -0.66 $\pm$ 0.01 & 0.15 $\pm$ 0.00 \\
 & & wo. last term & 0.47 $\pm$ 0.01 & -0.66 $\pm$ 0.01 & 0.03 $\pm$ 0.00 \\
\midrule
\multirow{2}{*}{solar} &
\multirow{2}{*}{1066/10}
 & w. last term & 0.93 $\pm$ 0.07 & -1.57 $\pm$ 0.20 & 0.07 $\pm$ 0.00 \\
 & & wo. last term & 0.93 $\pm$ 0.07 & -1.56 $\pm$ 0.20 & 0.03 $\pm$ 0.00 \\
\midrule
\multirow{2}{*}{pumadyn32nm} &
\multirow{2}{*}{8192/32}
 & w. last term & 1.00 $\pm$ 0.01 & -1.42 $\pm$ 0.01 & 21.12 $\pm$ 0.06 \\
 & & wo. last term & 1.00 $\pm$ 0.01 & -1.42 $\pm$ 0.01 & 0.05 $\pm$ 0.00 \\
\midrule
\bottomrule
\end{tabular}
\caption{RMSE, log-likelihood, and run time for two variants of predictive variance computation. }
\label{tab:results}
\end{table}

\begin{figure*}[t]
    \centering
    \includegraphics[width=\linewidth]{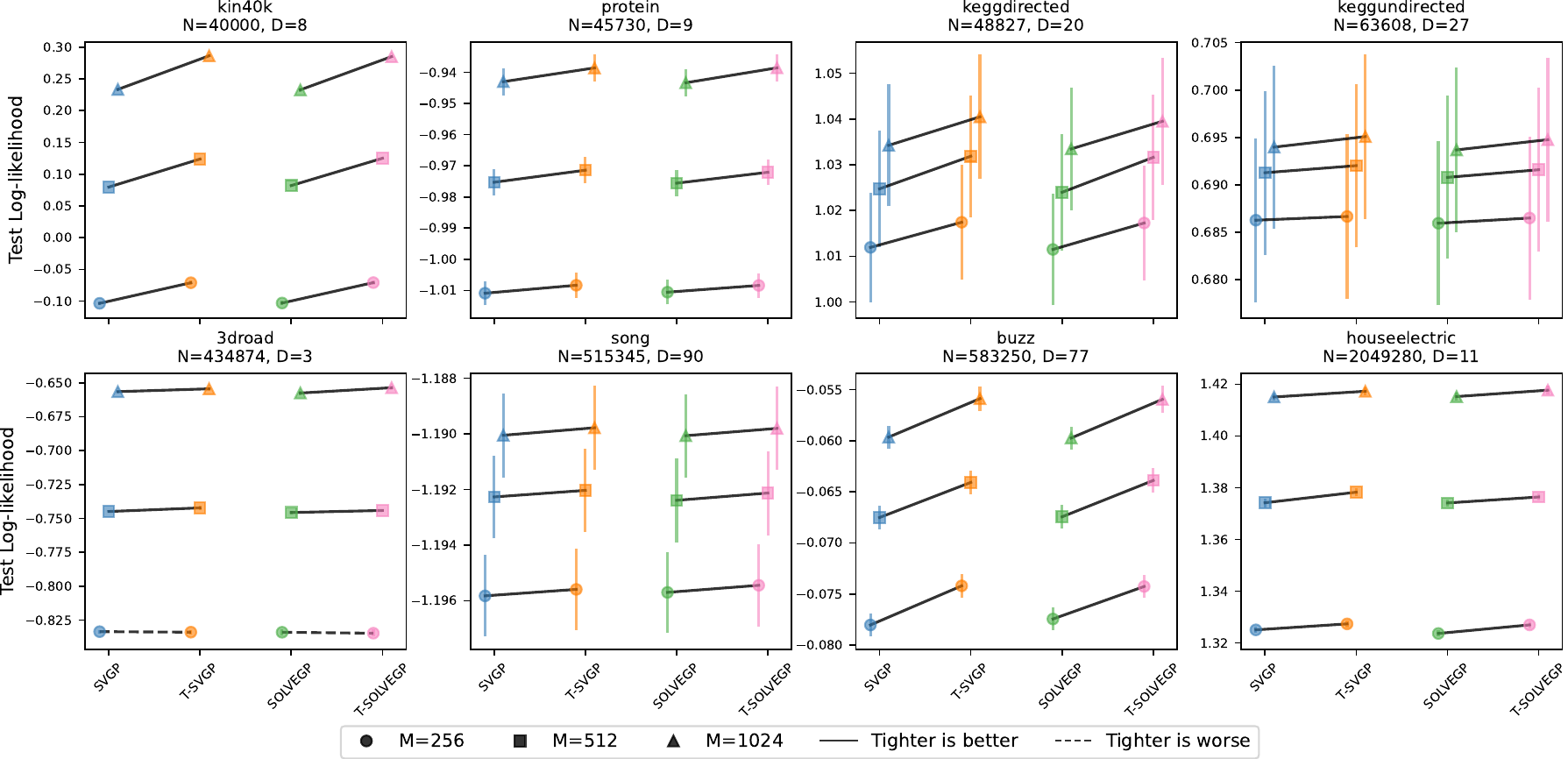}
    \caption{Test log-likelihood for various sparse approximations on eight regression datasets and various numbers of pseudo-points. For SOLVEGP and T-SOLVEGP, M is evenly split for $\vu$ and $\vv$. Higher is better. Best viewed in colour.}
    \label{fig:reg_nll}
\end{figure*}

\subsection{Large-scale regression benchmarks}
\label{sec:exp_reg}
\begin{figure}[!t]
    \centering
    \includegraphics[width=0.6\linewidth]{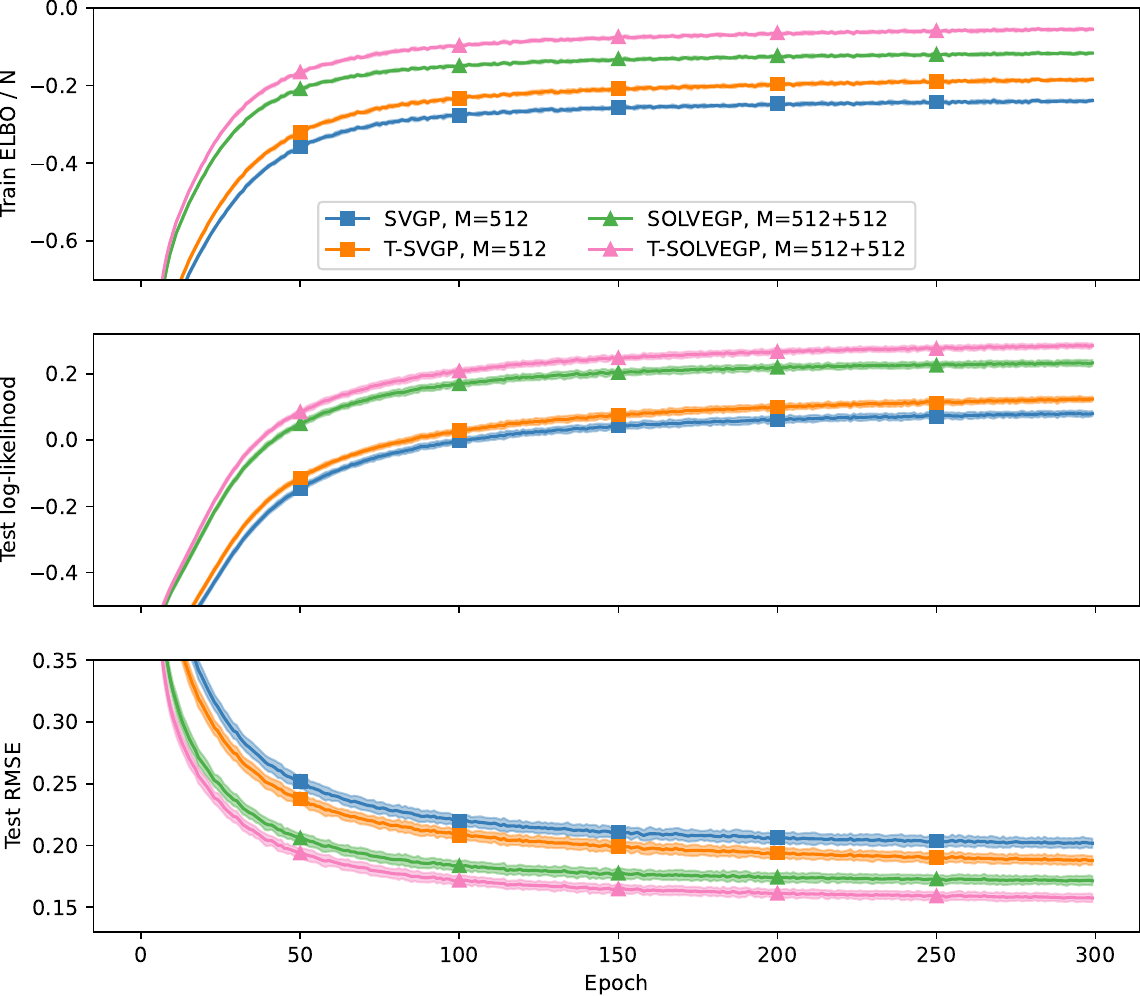}
    \caption{Variational bound and test performance for various approximations trained on the kin40k dataset. Best viewed in colour.}
    \label{fig:kin40k_traces}
\end{figure}

We next compare four methods, SVGP, T-SVGP, SOLVEGP, and the SOLVEGP variant in \cref{eq:tighter_solvegp} [T-SOLVEGP], across three inducing-point configurations ($M=256, 512, 1024$), on eight medium to large regression datasets. The datasets range from 40K to 2M data points with varying input dimensionalities \citep{yang15b}. We use the Matern-3/2 kernel and repeat each experiment 10 times, each employing a random train/test split. The comparison results are shown in \cref{fig:reg_nll,fig:reg_rmse,fig:reg_elbo}.
We note that (i) both T-SVGP and T-SOLVEGP consistently match or slightly outperform (on 5/8 datasets), or significantly outperform (on 3/8 datasets) their base counterparts, (ii) the performance improvement is also consistent across various inducing-point configurations, (iii) the improvements (on 3/8 datasets) are also consistent across training runs and iterations, as shown in \cref{fig:kin40k_traces} for the kin40k dataset, and (iv) SVGP and T-SVGP (and similarly SOLVEGP and T-SOLVEGP) have almost identical run time so the improvements here do not come at any cost.

\subsection{MNIST classification}
\label{sec:exp_mnist}
\begin{figure*}[!t]
    \centering
    \includegraphics[width=\linewidth]{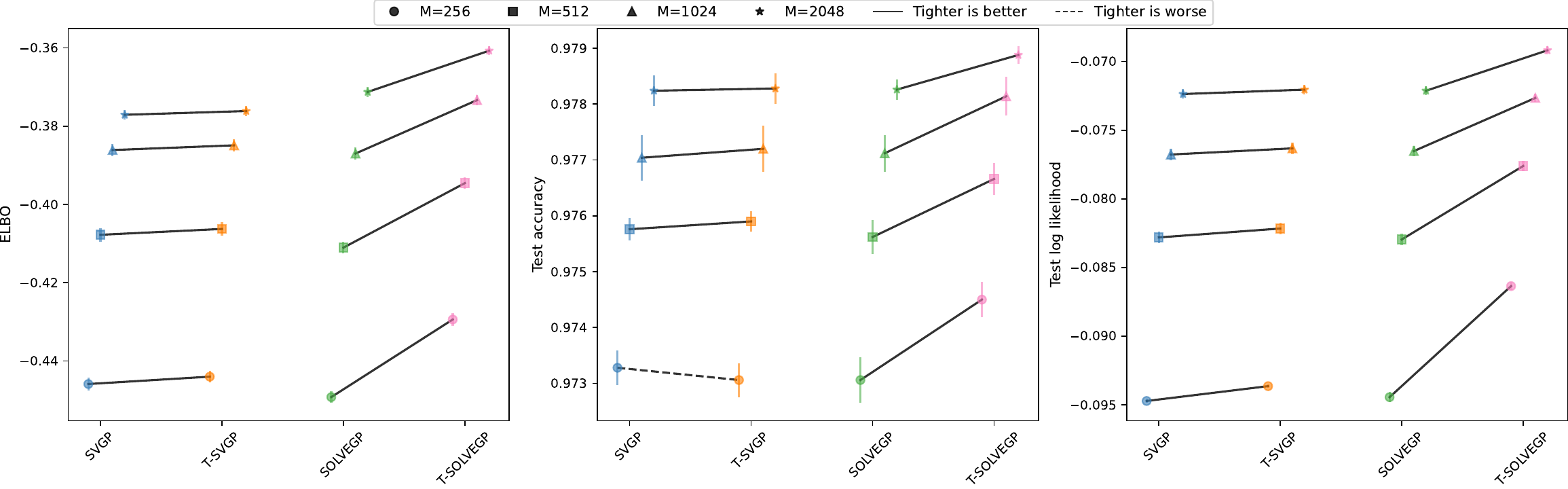}
    \caption{Log marginal likelihood approximations and test performance on the MNIST 10-way classification task. Best viewed in colour.}
    \label{fig:mnist_results}
\end{figure*}

To evaluate the performance of the proposed approximation on non-Gaussian likelihoods, we run an experiment on the MNIST digit classification task with 256, 512, 1024, and 2048 inducing points, using the SVGP, T-SVGP, SOLVEGP, and T-SOLVEGP variational objectives. \Cref{fig:mnist_results} shows that both the proposed approximations achieve substantial performance gains in all metrics compared to their base versions. 

\subsection{GPLVM on the oil flow dataset}
\begin{figure}[!t]
    \centering
    \includegraphics[width=0.6\linewidth]{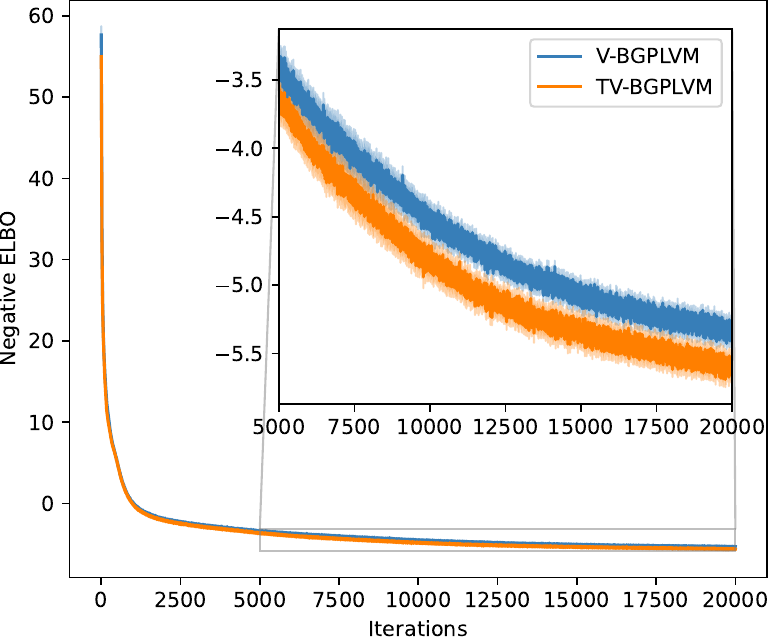}
    \caption{Optimisation traces for variational Bayesian GPLVM on the oil flow dataset. Best viewed in colour.}
    \label{fig:oil_flow}
\end{figure}

Finally, we demonstrate the proposed method's applicability to latent variable models through experiments with Bayesian GPLVM on the oil flow dataset.  The multi-phase oil flow dataset consists of 1000, 12-dimensional data points belonging to three classes which correspond to the different phases of oil flow in a pipeline \citep{bishop1993analysis}. \Cref{fig:oil_flow} compares the standard variational BGPLVM \citep{damianou16a,lalchand22a} [V-BGPLVM] against the proposed approximation in \cref{sec:lvm} [TV-BGPLVM]. The optimisation trajectories show that TV-BGPLVM achieves a lower final negative ELBO (roughly $-5.5$ versus $-5.2$), indicating a more accurate posterior approximation.

\section{Summary}
We build upon the standard sparse variational Gaussian process (SVGP) approximate posterior distribution through a simple modification to the conditional GP prior distribution at observed inputs. Using our proposed posterior approximation, we derive a collapsed bound which improves upon existing SVGP lower bounds to the log marginal likelihood, and an uncollapsed form which facilitates its application with non-Gaussian likelihoods and is compatible with stochastic mini-batch optimisation. Furthermore, we show how our approach can be used to improve non-standard SVGP posterior approximations, such as SOLVE-GP \citep{shi2020sparse}.

Our empirical results demonstrate consistent improvements in both predictive performance and log marginal likelihood estimates across diverse applications, including regression, classification, and latent variables modelling tasks. The proposed posterior approximations can be easily applied to other settings such as deep GPs and convolutional GPs \citep{van2017convolutional,blomqvist2020deep,sun2021scalable,bui16,salimbeni2017doubly}.

%% file: appendix_text.tex
\section{An even tighter but expensive approximation}
We consider a more general form for the conditional covariance of $q(\vf | \vu)$ as follows:
\begin{align}
    q(f) &= p(f_{\neq \vf, \vu} | \vf, \vu) q(\vf | \vu) q(\vu), \nonumber\\
    q(\vf | \vu) &= \gN(\vf; \kfu\kuuinv\vu; \rmC),\nonumber
\end{align}
Again, we can also obtain the optimal form for $ q(u) \propto p(\vu) \gN(\vy; \kfu\kuuinv\vu, \sigma^2\rmI)$, leading to the following collapsed bound
\begin{align}
    \gF_6(\theta) &= \vc - \frac{1}{2} \vy^\intercal (\qff + \sigma^2\rmI)^{-1} \vy - \frac{1}{2} \log |\qff + \sigma^2\rmI| - \frac{1}{2}\tr[(\sigma^{-2} \rmI + \dff^{-1})\rmC] - \frac{1}{2} \log |\rmC^{-1}\dff|. \nonumber
\end{align}
We can derive the optimal $\rmC$, $\rmC^{-1} = \dff^{-1} + \sigma^{-2}\rmI$ and the bound becomes:
\begin{align}
    \gF_8(\theta) 
    &= \vc - \frac{1}{2} \vy^\intercal (\qff + \sigma^2\rmI)^{-1} \vy - \frac{1}{2} \log |\qff + \sigma^2\rmI| - \frac{1}{2} \log |\rmI + \sigma^{-2}\dff| \nonumber\\
    &= \vc - \frac{1}{2} \vy^\intercal (\qff + \sigma^2\rmI)^{-1} \vy - \frac{1}{2} \log |\qff + \sigma^2\rmI| - \frac{1}{2} \sum_n \log (1 + \sigma^{-2}\lambda_n(\dff)|,\nonumber
\end{align}
where $\lambda_n(\rmX)$ is the $n$-th eigenvalue of $\rmX$. The bound above is as expensive as the original log marginal likelihood.

\section{Exploring alternative parameterisations for the conditional posterior}
Instead of the general $\rmC$ as above or the form considered in the main text $\rmC = \dff^{1/2}\rmM \dff^{\top/2}$, we consider two other parameterisations that might allow efficient collapsed/un-collapsed bounds and predictions. We first rewrite the uncollapsed bound and the predictive mean and variance here for clarity,
\begin{align*}
    \gF_{\textrm{uncollapsed}}
        &= - \kl [q(\vu) || p(\vu)] - \int_\vu q(\vu) \kl [q(\vf | \vu) || p(\vf | \vu)] +  \sum_{n} \int_{\vu, f(x_n)} q(\vu) q(f({x_n}) | \vu) \log p(y_n | f({x_n})), \\
    m_* &= \ksu \kuuinv \rvm_\vu, \\
    v_* &= \kss - \ksu \kuuinv \kus + \ksu \kuuinv \rmS_\vu \kuuinv \kus - (\ksf - \qsf) (\dff - \rmC) (\kfs- \qfs),
\end{align*}

We first consider $\rmC = \rmM^{1/2}\dff\rmM^{1/2}$. While this allows efficient exact predictive marginal distributions at training points, the middle term in the bound is costly to compute due to the presence of $\dff$: 
\begin{align}
    - \int_\vu q(\vu) \kl [q(\vf | \vu) || p(\vf | \vu)]
    = - \frac{1}{2} \tr(\dff^{-1}\rmM^{1/2}\dff\rmM^{1/2}) + \frac{1}{2}\log |\rmM| + \frac{N}{2}\nonumber.
\end{align}

Another special case of the parameterisation presented in the main text is $\rmC = m \dff$, i.e., a single $m$ is shared across all training points. This conveniently leads to tractable exact predictive variances at training points, $v_n = m d_n + \mathbf{k}_{f_n\vu} \kuuinv \rmS_\vu \kuuinv \mathbf{k}_{\vu f_n}$. The middle term in the bound can be simplified to,
\begin{align}
    - \int_\vu q(\vu) \kl [q(\vf | \vu) || p(\vf | \vu)]
    = \frac{N}{2}[1 + \log(m) - m] \nonumber.
\end{align}
In the regression case, this leads to  the optimal $m = \sigma^2 / (N^{-1}\sum_n d_n + \sigma^2)$ and the following collapsed bound,
\begin{align}
    \gF_9(\theta) =\vc - \frac{1}{2} \vy^\intercal (\qff + \sigma^2\rmI)^{-1} \vy - \frac{1}{2} \log |\qff + \sigma^2\rmI| - \frac{N}{2} \log \left( 1+\frac{\sum_n d_n}{N\sigma^2} \right).
\end{align}
This bound is looser than the collapsed bound in \cref{eq:tighter_collapsed}, due to the Jensen's inequality $\log (1 + \sum_n x_n / N) \geq N^{-1}\sum_n \log(1 + x_n)$.

\section{Additional experimental results}
\subsection{Large-scale regression benchmarks}
\begin{figure*}[!ht]
    \centering
    \includegraphics[width=\linewidth]{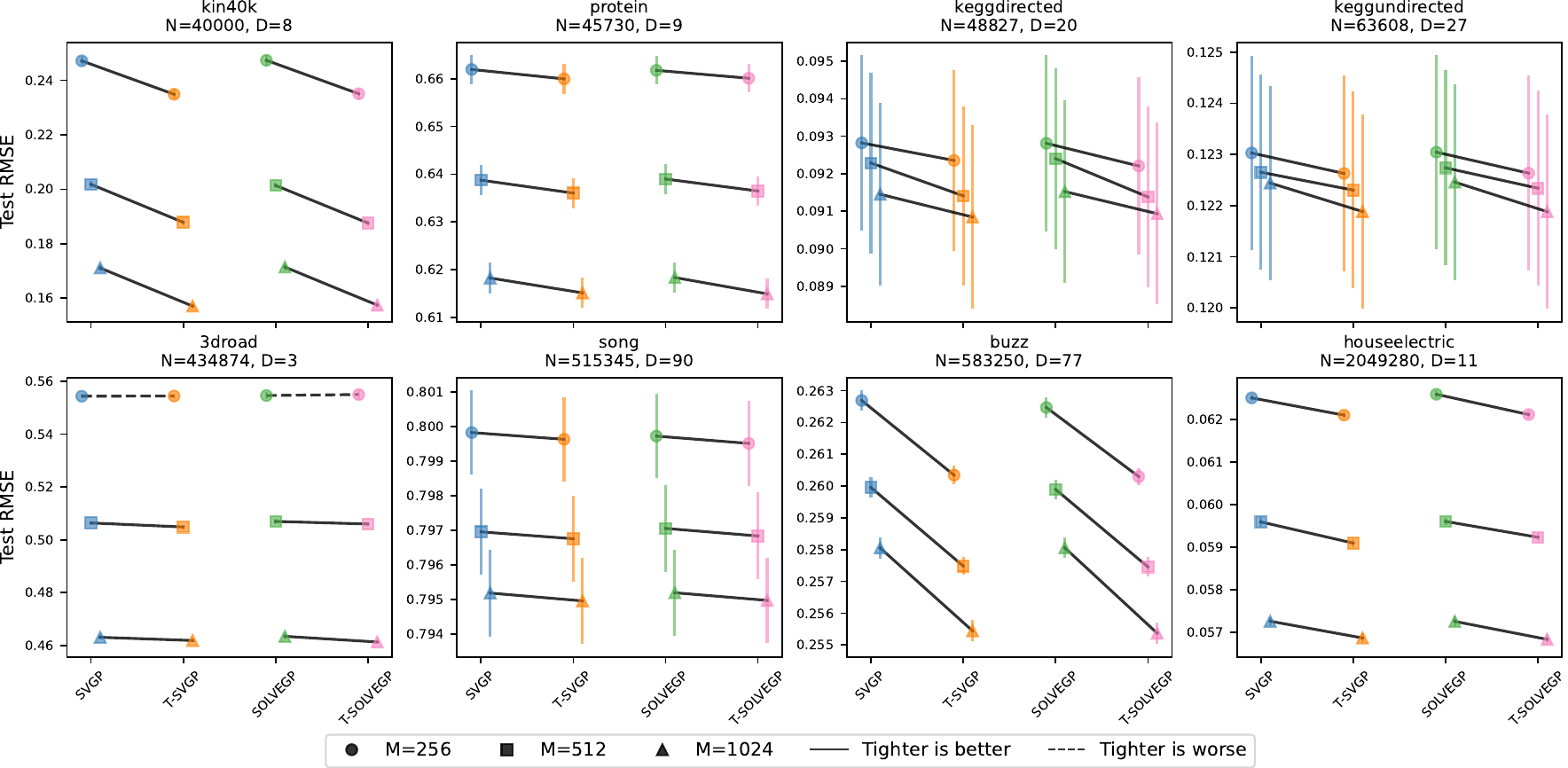}
    \caption{Test root mean squared errors (RMSE) for various sparse approximations on eight regression datasets and various numbers of pseudo-points. For SOLVEGP and T-SOLVEGP, M is evenly split for $\vu$ and $\vv$. Lower is better. Best viewed in colour.}
    \label{fig:reg_rmse}
\end{figure*}

\begin{figure*}[!ht]
    \centering
    \includegraphics[width=\linewidth]{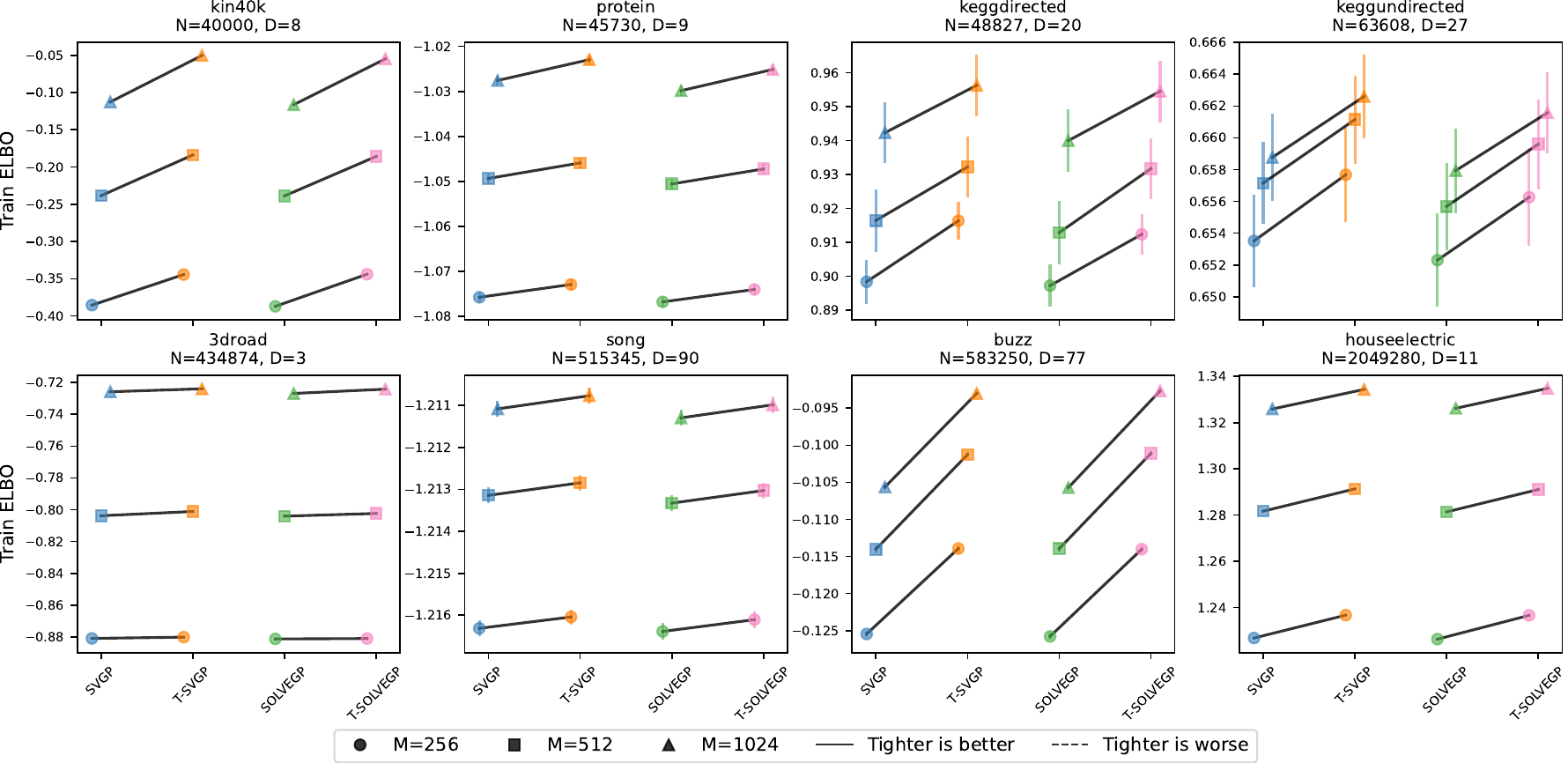}
    \caption{Log marginal likelihood approximations (ELBO) for various sparse approximations on eight regression datasets and various numbers of pseudo-points. For SOLVEGP and T-SOLVEGP, M is evenly split for $\vu$ and $\vv$. Higher is better. Best viewed in colour.}
    \label{fig:reg_elbo}
\end{figure*}